\documentclass{article} 
\usepackage[final]{colm2026_conference}

\usepackage{microtype}
\usepackage{hyperref}
\usepackage{url}
\usepackage{booktabs}

\usepackage{amsmath}
\usepackage{amssymb}
\usepackage{mathtools}
\usepackage{amsthm}
\usepackage{makecell}
\usepackage{graphicx}
\usepackage{subcaption}
\usepackage{xcolor}
\usepackage{multirow}
\usepackage{multicol}
\usepackage{pifont}
\usepackage[misc]{ifsym}

\newcommand{\circleone}{\ding{172}}
\newcommand{\circletwo}{\ding{173}}
\newcommand{\circlethree}{\ding{174}}
\newcommand{\circlefour}{\ding{175}}

\usepackage[capitalize,noabbrev]{cleveref}

\definecolor{darkblue}{rgb}{0, 0, 0.5}
\hypersetup{colorlinks=true, citecolor=mid, linkcolor=primary, urlcolor=mid}

\newcommand{\codename}{MeKi}
\newcommand{\eg}{\textit{e.g.}}

\renewcommand{\cite}[1]{\citep{#1}}

\title{\centering \codename: Memory-based Expert Knowledge Injection for  Efficient LLM Scaling }

\author{
\mbox{
Ning Ding$^{1\dagger}$, Fangcheng Liu$^{1\dagger}$, Kyungrae Kim$^2$, Linji Hao$^1$, Kyeng-Hun Lee$^2$,
}\\
Hyeonmok Ko$^2$, and Yehui Tang$^{1{~\textrm{\Letter}}}$ \\
$^1$ Samsung Research, Beijing, China \quad\quad
$^2$ Samsung Research, South Korea\\
\texttt{\{ning1.ding, yehui.tang\}}@samsung.com \\
$^\dagger$~Equal Contribution\quad\quad$^\textrm{\Letter}$~Corresponding Author
}

\begin{document}

\maketitle
\begin{abstract}
Scaling Large Language Models (LLMs) typically relies on increasing the number of parameters or test-time computations to boost performance. However, these strategies are impractical for edge device deployment due to limited RAM and NPU resources. Despite hardware constraints, deploying performant LLM on edge devices such as smartphone remains crucial for user experience. To address this, we propose \textbf{\codename}~(\textbf{M}emory-based \textbf{E}xpert \textbf{K}nowledge \textbf{I}njection), a novel system that scales LLM capacity via storage space rather than FLOPs. \codename~equips each Transformer layer with token-level memory experts that injects pre-stored semantic knowledge into the generation process. To bridge the gap between training capacity and inference efficiency, we employ a re-parameterization strategy to fold parameter matrices used during training into a compact static lookup table. By offloading the knowledge to ROM, \codename~decouples model capacity from computational cost, introducing zero inference latency overhead. Extensive experiments demonstrate that \codename~significantly outperforms dense LLM baselines with identical inference speed, validating the effectiveness of memory-based scaling paradigm for on-device LLMs. Project homepage is at \url{https://github.com/ningding-o/MeKi}.
\end{abstract}

\begin{figure}[!h]
\centering    
\includegraphics[width=0.59\columnwidth]{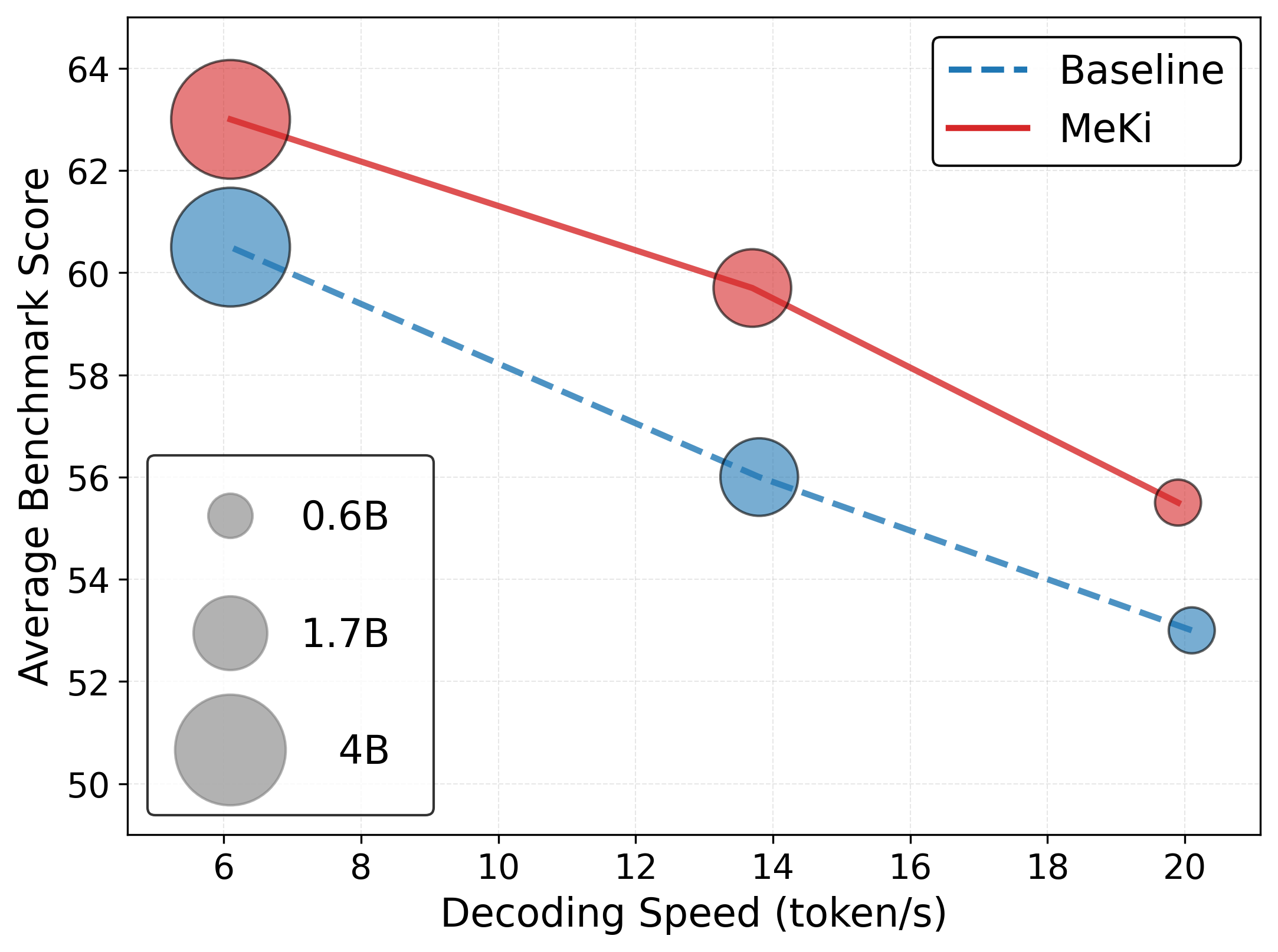}
  \vspace{-2mm}
  \caption{{Memory-based scaling for on-device deployment.} Scaling parameters of LLM makes inference much slower on NPU\protect\footnotemark, which motivates our methodology of memory-based scaling by offloading static knowledge into Read-Only Memory (ROM). 1.7B-\codename~achieves an average zero-shot benchmark score of 59.7, statistically rivaling the 4B dense model's score of 60.5, while maintaining a 2.26$\times$ advantage in decoding speed.
  }
  \label{fig:teaser}
\end{figure}
\footnotetext{~We measure the generation speed (token/s) on Qualcomm Snapdragon 8 Elite mobile platform with KV cache length being 10K.}

\section{Introduction}

The scaling law~\cite{kaplan2020scaling, hoffmann2022training} for Large Language Models (LLMs) has become the de-facto methodology to achieve increasingly higher performance among all the flagship models. To follow this trajectory, some choose to crawl and synthesize massive training corpora~\cite{raffel2020exploring,almazrouei2023falcon,penedo2024fineweb} for lengthened training stages. Alternatively, others choose to continuously increase the number of parameters within the Transformer architecture. This includes the previous SwiGLU~\cite{shazeer2020glu} method which expands the feed-forward networks (FFNs), as well as the prevalently state-of-the-art Mixture-of-Experts (MoE)~\cite{shazeer2017outrageously,fedus2022switch,jiang2024mixtral} architecture which comprises multiple FFNs but only activates a sparse subset for each token. Recently, researchers find that scaling test-time FLOPs is also able to improve the performance. By exploring the reasoning path~\cite{wei2022chain}, verifying intermediate outputs~\cite{cobbe2021training}, or employing search-based decoding strategies~\cite{yao2023tree,snell2024scaling}, models can effectively trade latency for accuracy. This approach shifts the focus from training-time weight updates to dynamic test-time computation during the deployment stage.

While these compute-heavy scaling strategies have achieved remarkable success in data-center environments, they encounter significant obstacles when deployed to edge devices. Enlarging the parameters of dense model increases the floating-point operations (FLOPs), leading to unacceptable latency and power consumption on mobile hardware, to which users are usually very sensitive. \cref{fig:teaser} demonstrates the detailed inference speed of dense LLMs on the smartphone platform with on-device Qualcomm NPU~\cite{qualcomm_processor}. Although MoE architectures reduce per-token FLOPs by sparsely activating experts, this architecture introduces substantial latency overhead due to the frequent loading of large and disjoint expert weights. The memory access pattern of MoE would becomes the primary latency bottleneck on resource-constrained hardware such as mobile phone, where ROM-RAM and RAM-NPU bandwidth is far more limited than server-grade GPUs in data centers.

Consequently, scaling LLMs for edge deployment requires a new paradigm shifted away from compute-centric design. Unlike matrix multiplications which are expensive in computation, we observed that memory lookups (\eg~reading from ROM storage) are relatively cheap and energy-efficient on modern mobile System-on-Chips (SoC). Besides, the ROM bandwidth is largely idle during model inference.This observation motivates a critical question:
\begin{quote}
\vspace{-1mm}
\textit{Can we scale up the model capacity by using the storage space, without increasing the latency and FLOPs during inference?}
\end{quote}
\vspace{-1mm}
To answer this, we introduce \textbf{\codename} (\textbf{M}emory-based \textbf{E}xpert \textbf{K}nowledge \textbf{I}njection), a novel LLM architecture which decouples the model capacity from computational cost. Just like MoE model retrieves specialized experts for different tokens, \codename~retrieves token-level dedicated knowledge vectors from a massive memory bank in every layer, injecting the learned knowledge into the hidden states.

To increase model capacity while maintaining efficiency, \codename~has different structures for training and inference. During the training stage, \codename~employs embedding-based memories along with non-linear projections to learn token-level expert knowledge representations. These features are fused with the hidden state via a low-rank gating mechanism to inject useful information into the hidden sequence. 
After training, we use re-parameterization technique to merge the complex projections into representations stored in the memory bank, which is offloaded to ROM. This turns the heavy online computation into a compact static lookup operation during inference. The \codename~system operates in parallel with the Transformer’s FFN module, which allows for implicit layer expansion nearly without extra FLOPs overhead.

The contributions of this work are summarized as follows:
\begin{itemize}
\vspace{-2mm}
\item We propose \textbf{\codename}, a memory-centric scaling method, which decouples the LLM capacity growth from computation. We treat the layer-wise memory as a collection of token-level experts that carry rich prior knowledge. We design a efficient fusion architecture parallel to the FFN module, using knowledge injection to expand model capacity.
\item We introduce a re-parameterization strategy that balance the training-time capacity and test-time efficiency. Our strategy allows the use of complex non-linear projections during training to maximize feature representability, while these projections are then merged into a static embedding table for zero-cost inference.
\item We validate the effectiveness of the proposed \codename~architecture across the scales of 0.6B, 1.7B, and 4B parameters. Our method outperforms the dense baseline across 10 widely-used benchmarks.
\item We test the inference latency of \codename~architecture and its baseline counterpart on Qualcomm Snapdragon android hardware. Our method maintains the same inference speed with significant performance gains.
\end{itemize}

\section{Related Works}

\paragraph{Evolution of LLM Architectures.} The trajectory of LLM development~\cite{touvron2023llama, achiam2023gpt} has reached a critical turning point. While scaling laws continue to drive performance gains in massive data centers, the challenge of deploying these models on edge devices remains significantly underexplored. The Mixture-of-Experts (MoE) paradigm~\cite{shazeer2017outrageously, fedus2022switch, liu2024deepseek} scales model capacity by sparsely activating parameter subsets. However, MoE introduces substantial inference overhead on edge hardware due to dynamic routing latency and memory fragmentation. In contrast, \codename~proposed in this paper eliminates online routing costs and enables seamless prefetching, making it more suitable for resource-constrained environments.

\paragraph{New Scaling Paradigm Beyond RAM.} An upsurging research direction aims to scale model performance by leveraging external storage rather than increasing active memory (RAM) or FLOPs: \emph{1) Retrieval-Augmented Performance Scaling}~\cite{khandelwal2019generalization,borgeaud2022improving,asai2024self,huang2024ultra} that augments LLMs by retrieving information from massive corpora. \emph{2) Memory-Centric Capacity Expansion.} \citet{zeng2023lookupffn}~proposed LookupFFN which explored replacing the whole FFNs with lookup tables, and \citet{sadhukhan2026stem}~used lookup tables to replace part of the FFN. \citet{ding2024memoryformer}~further replaced all linear layers in model with lookup tables. Per-Layer Embedding~\cite{gemma_3n_2025} (PLE) utilized tokenID-indexed embedding memory to expand the model depth, but is prone to forming information bottlenecks in the forward pass. The proposed \codename~advances beyond PLE by parallelizing embedding memory with FFN to serve as a capacity extender.
Concurrent to our work, Engram~\cite{cheng2026conditional} utilizes $N$-gram statistics for phrase-level caching. Unlike Engram which relies on online hashing for knowledge retrieval, \codename~utilizes a low-rank gating mechanism to dynamically augment hidden states with token-level expert knowledge, providing superior contextual adaptation and lower latency for edge deployment.

\begin{figure*}[t]
\centering
\includegraphics[width=0.8\textwidth]{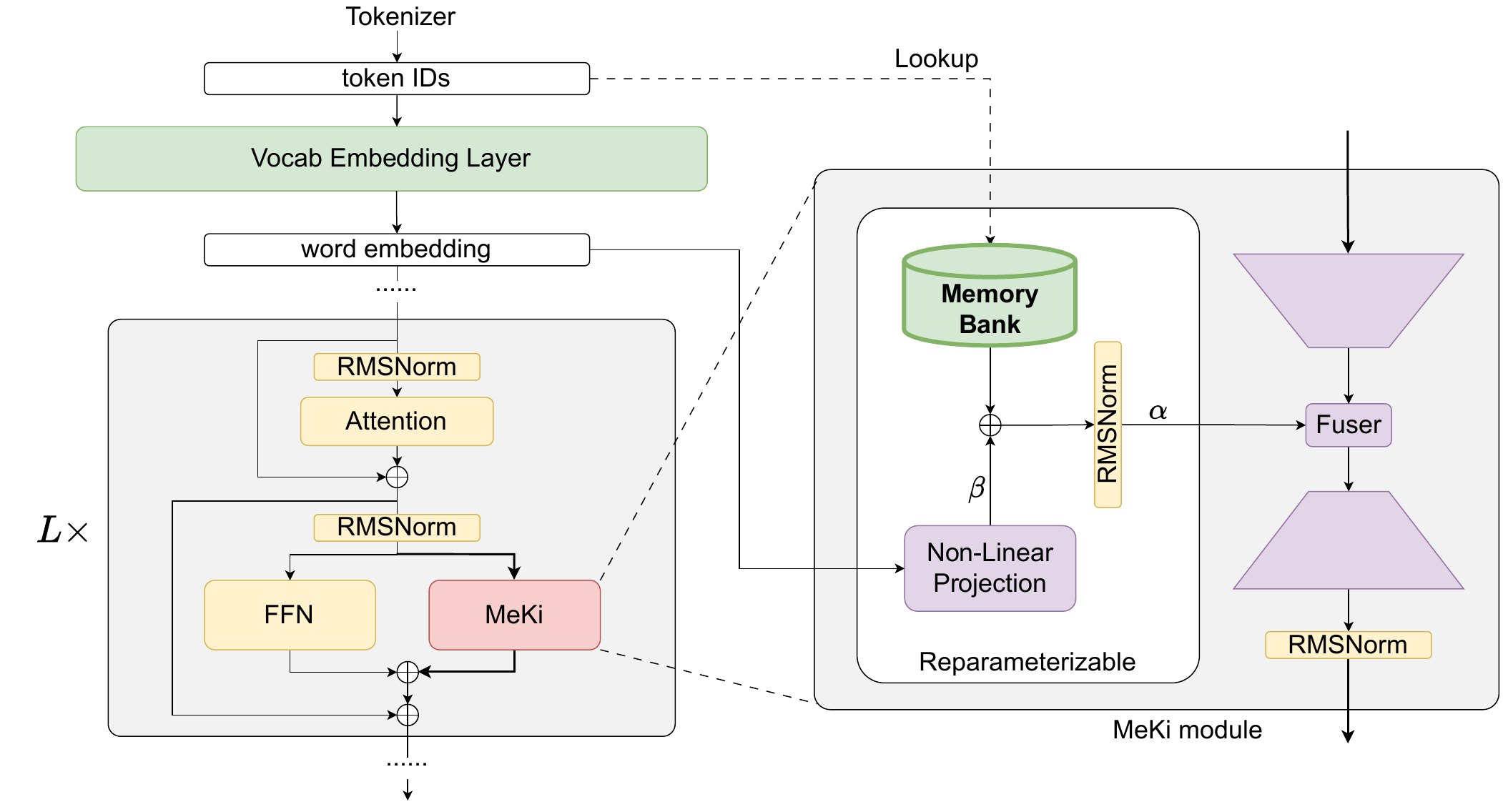}
\caption{General overview of the proposed \textbf{\codename}~architecture. After training, all the re-parameterizable parts are merged into one static memory table to avoid FLOPs overhead.}
\label{fig:architecture}
\end{figure*}

\section{Architecture}
\label{sec:architecture}

\subsection{Overview}
\label{ssec:overview}

In this section, we detail the architecture of the proposed \textbf{M}emory-based \textbf{E}xpert \textbf{K}nowledge \textbf{I}njection (\codename) system. We first provide a high-level overview of the framework, followed by a detailed formulation of each component. \cref{ssec:expert} details the composition of the Token-Level Expert during training stage.
\cref{ssec:fusion} details how to inject the retrieved knowledge into hidden state via a low-rank feature fusion mechanism. \cref{ssec:reparam} proposes a re-parameterization strategy for inference optimization.

The core motivation of \codename~is to decouple the model capacity from computational cost by shifting the burden from floating-point matrix multiplications (FLOPs) to high-density memory lookups. 
 \cref{fig:architecture} illustrates the integration of \codename~within a standard Transformer layer. \codename~operates as a parallel branch to the standard Feed-Forward Network (FFN). 
For the $l$-th layer, \codename~retrieves a specific knowledge vector for every token from a layer-specific memory bank. This retrieval is then used to modulate the input hidden state to generate context-aware expert knowledge. The branch then projects the token knowledge back to the model dimension and adds it to the residual stream. Crucially, the architecture utilizes different parameter configurations for training and inference. During training, it employs complex non-linear projections to learn rich representations. Before deployment, a re-parameterization technique folds these complex computations into the memory bank weights, which guarantees that the inference latency in not compromised.

\subsection{Memory-based Token-Level Expert}
\label{ssec:expert}

The \codename~module conceptualizes the embedding memory as a sparse token-level expert. Unlike Mixture-of-Experts (MoE) models that activate different network weights based on routing mechanism, \codename~retrieves expert knowledge directly from a Read-Only Memory (ROM) structure, ensuring that for any input token, only the specific vector corresponding to its ID is activated.

Given an input sequence with corresponding input token IDs $\mathbf{x} \in \{1, \dots, |\mathcal{V}|\}^T$, $T$ is the sequence length and $|\mathcal{V}|$ is the vocabulary size. For the $l$-th transformer layer, we define a layer-specific memory embedding matrix $\mathbf{M}^l \in \mathbb{R}^{|\mathcal{V}| \times d_{mem}}$, where $d_{mem}$ is the memory dimension (\eg, 128 or 256) and is typically much smaller than $d_{model}$.
Given an input token ID $x_t$, the static retrieved memory vector $\mathbf{m}^l_{static}(x_t)$ is obtained via a standard embedding lookup:
\begin{equation}
    \mathbf{m}^l_{static}(x_t) = \mathbf{M}^l[x_t].
\end{equation}

To further enhance the representation ability of the expert without increasing inference cost, we introduce a dynamic projection component during training. This component utilizes a non-linear projection function $\mathbf{G}^l(\cdot)$ (implemented as an SwiGLU in practice) acting on the global word embedding $E_{global} \in \mathbb{R}^{|\mathcal{V}| \times d_{model}}$. This allows the model to better synthesize layer-specific knowledge features from the shared semantic space:
\begin{equation}
    \mathbf{m}^l_{dyn}(x_t) = \mathbf{G}^l(E_{global}[x_t]).
\end{equation}

The final expert vector $e^l_t$ for the $t$-th token at layer $l$ is constructed by gathering the static prior knowledge and the dynamic projection. This fusion is stabilized by a normalization two learnable scalar $\alpha^l$ and $\beta^l$:
\begin{equation}
\mathbf{e}^l_t = \alpha^l \cdot \text{RMSNorm}\left( \mathbf{m}^l_{static}(x_t) +\beta^l \cdot  \mathbf{m}^l_{dyn}(x_t) \right).
\label{eq:memory_type}
\end{equation}

Here, $\beta^l$ allows \codename~to learn a layer-wise combining ratio between the static learned knowledge and the dynamic projected feature, and $\alpha^l$ controls the layer-specific knowledge injection magnitude.

\subsection{Knowledge Injection}
\label{ssec:fusion}

Once the expert vector $\mathbf{e}^l_t$ is retrieved, it must be injected into the Transformer's hidden state flow. A critical design choice in \codename~is the fusion mechanism, which modulates how the retrieved expert knowledge $\mathbf{e}^l_t$ interacts with the $t$-th token hidden state $\mathbf{h}^l_t \in \mathbb{R}^{d_{model}}$. It's worth noting that, we let \codename~block share the same input hidden state $\mathbf{H}$ as the FFN block, where $\mathbf{H} = \{\mathbf{h}^l_1 ,\cdots, \mathbf{h}^l_T \}$ is the output of the preceding normalization layer.

We employ an additive gated fusion mechanism, which offers superior optimization stability compared to multiplicative gating. Specifically, we first generate a gate signal $\mathbf{g}_t$ by using the hidden state via a low-rank linear projection followed by a sigmoid activation:
\begin{equation}
    \mathbf{g}^l_t = \sigma(\mathbf{W}^{l}_{gate} \mathbf{h}^l_t),  ~~\mathbf{W}^{l}_{gate} \in \mathbb{R}^{d_{mem} \times d_{model}}.
\end{equation}

The expert vector $\mathbf{e}^l_t$ is then modulated by adding this gate signal, implying that $\mathbf{g}^l_t$ serves as a contextualized channel-wise offset to adjust the expert knowledge:
\begin{equation}
    \mathbf{v}^l_t = \mathbf{e}^l_t + \mathbf{g}^l_t.
\end{equation}

The modulated vector $\mathbf{v}_t$ resides in the memory dimension $d_{mem}$. To inject it back into the main residual stream of the Transformer layer, we project it up to the hidden dimension $d_{model}$ using another linear projection:
\begin{equation}
    \mathbf{y}^l_t = \text{RMSNorm}\left( \mathbf{W}^{l}_{out} \mathbf{v}^l_t \right),  ~~\mathbf{W}^{l}_{out} \in \mathbb{R}^{d_{model} \times d_{mem}}.
\end{equation}

The final output of the \codename~module $\mathbf{y}^l_t$ is then added to the residual connection of the Transformer layer. If we denote the FFN block as $\mathbf{F}(\cdot)$ and the input sequence as $\mathbf{H}$, the output of the $l$-th Transformer layer is updated as:
\begin{align}
\mathbf{H} &= \text{RMSNorm}_{\text{pre}\_\text{ffn}}(\mathbf{A})  \\
\mathbf{H}' &= \mathbf{F}(\mathbf{H}) + \mathbf{\codename}(\mathbf{H}) + \mathbf{A},
\end{align}

where $\mathbf{A}$ is the output sequence of the residual-add op after Attention module. This parallel structure allows \codename~to function as an implicit width expansion of the Transformer layer, enhancing capacity without affecting the original connection path of the FFN module.

\subsection{Re-Parameterization for Faster Inference}
\label{ssec:reparam}

While the non-linear projection $\mathbf{G}^l(\cdot)$ in the \codename~module significantly boosts the model's learning capacity, it also introduces additional FLOPs during the forward pass. To achieve the goal of zero-cost inference, we propose a re-parameterization technique that merges the dynamic projection $\mathbf{m}^l_{dyn}(\cdot)$ into the embedding table $\mathbf{M}^l$ of static memory.

Observing that the dynamic projection $\mathbf{G}^l(\cdot)$ operates solely on the global word embedding weights $E_{global}$, which are static constants after training, we can pre-compute the output of this projection. Specifically, we define a new embedding table $\tilde{\mathbf{M}}^l$ as follows:
\begin{align}
    \tilde{\mathbf{M}}^l &= \text{Re-parameterize}\left( \mathbf{M}^l, E_{global} ,\textbf{G}^l\right) \\
    &= \alpha^l \cdot \text{RMSNorm}\left( \mathbf{M}^l + \beta^l \cdot \textbf{G}^l(E_{global}) \right).
\end{align}

By replacing the original memory weight $\mathbf{M}^l$ with $\tilde{\mathbf{M}}^l$, the derivation of token-level expert vector at inference time is simplified drastically. The term $\mathbf{G}^l(E_{global}[x_t])$ gets absorbed into the new lookup table, eliminating the need for any matrix multiplication during the generation of expert knowledge vector $\mathbf{e}^l_t$.

Consequently, the knowledge retrieval at inference-time becomes:
\begin{equation}
    \tilde{\mathbf{e}}^l_t = \tilde{\mathbf{M}}^l[x_t],~~ \tilde{\mathbf{M}}^l \in \mathbb{R}^{|\mathcal{V}| \times d_{mem}},
\end{equation}
followed by the lightweight fusion step. This fusion step involves only two small matrix multiplications of size $d_{mem} \times d_{model}$ and an addition. Therefore, the inference process of \codename~is:
\begin{align}
    \mathbf{g}^l_t &= \sigma(\mathbf{W}^{l}_{gate} \mathbf{h}^l_t), \\
    \mathbf{v}^l_t &= \tilde{\mathbf{e}}^l_t + \mathbf{g}^l_t, \\
    \mathbf{y}^l_t &= \text{RMSNorm}\left( \mathbf{W}^{l}_{out} \mathbf{v}^l_t \right)
\end{align}

\subsection{Computational Complexity Analysis}

We demonstrate the efficiency of the proposed \codename~by analyzing the FLOPs statistic per layer per token.
\paragraph{\textbf{Training FLOPs of \codename.} }The non-linear projection $\mathbf{G}^l$ which generates dynamic knowledge feature $\mathbf{m}^l_{dyn}(x_t)$ is implemented as an SwiGLU with intermediate size $\frac{d_{model}}{2}$. The FLOPs of $\mathbf{G}^l$ per token are $O(d_{model}^2+ \frac{d_{model}}{2} \cdot d_{mem})$. After considering the two low-rank gating projections $\mathbf{W}^{l}_{gate}$ and $\mathbf{W}^{l}_{out}$, the per-token FLOPs of \codename~module are $O(d_{model}^2+ \frac{5}{2}d_{model} \cdot d_{mem})$ during training stage.
\vspace{-3mm}
\paragraph{\textbf{Re-Parameterized Inference FLOPs.}} After the heavy projection $\mathbf{G}^l$ is removed, the remaining operations are the embedding lookup (negligible I/O cost) and the low-rank gating projections with FLOPs of $O(d_{model} \cdot d_{mem})$ level. Since the memory dimension $d_{mem} \ll d_{model}$ (\eg~128 vs 2048), the re-parameterized inference effectively shifts the complexity from matrix multiplication to memory access bandwidth.
\vspace{-3mm}
\paragraph{\textbf{On-device Circumstance.}} For a 28-layer model with $d_{mem}=256$, the memory weights to be moved from ROM is only $d_{mem}\times L=7168$, which corresponds to 14KB per token in float16 format.
On modern mobile SoC with NPU, embedding tables are typically cached in high-speed memory where ROM bandwidth is abundant (\eg~ UFS-4.0 delivers 4.2GB/s reading speed~\cite{samsung_ufs}). Therefore, the model obtained after re-parameterization has almost the same token generation latency compared to the baseline model without \codename.

\section{Experiments}
\label{sec:experiments}

In this section, we evaluate the proposed architecture to demonstrate that \codename~effectively decouples parametric memory from computational cost, achieving superior performance on comprehensive benchmarks without incurring inference latency overhead.

\subsection{Experimental Setup}
\label{sec: setup}

\paragraph{Datasets and Baselines.} 
We conduct pre-training on the \emph{FineWeb-Edu-Dedup} dataset ~\cite{benallal2024smollmcorpus}, which comprises high-quality educational contents collected from internet. We randomly sampled 50 billion tokens from this dataset and use the same 50B-subset to pre-train all the models in this paper for fair comparisons. The models are implemented using the Megatron-LM framework~\cite{megatron-lm}. We adopt the architecture of Qwen3-0.6B, -1.7B, and -4B~\cite{qwen3technicalreport} as our primary dense baselines. 

\paragraph{Training Details.}
All models are trained from scratch with the AdamW optimizer ($\beta_1=0.9, \beta_2=0.95$), using BFloat16 mixed-precision. To ensure training stability, we apply a weight decay of 0.1 and employ gradient clipping with a global norm threshold of 1.0. We use a cosine learning rate schedule with a warm-up phase of 500 steps. The peak and minimum learning rate is set to $4.0 \times 10^{-4}$ and $2.0 \times 10^{-4}$, respectively. We use the global batch size of 256 and the sequence length of 4,096. Each training step contains 1M tokens. All the models are trained for 50,000 steps. Detailed architecture and hyperparameter configurations are summarized in \cref{tab:hyper-llama-opt}.

\paragraph{Evaluation Benchmarks.} To evaluate our model's performance, we conduct evaluations across ten recognized benchmarks, categorized by required reasoning capabilities following LLaMA~\cite{touvron2023llama}: \emph{1) Science \& Knowledge}: ARC-E/C~\cite{clark2018think}, OBQA~\cite{mihaylov2018can}, and SciQ~\cite{welbl2017crowdsourcing} are used to test factual retrieval and scientific reasoning; \emph{2) Commonsense \& Causality}: PIQA~\cite{bisk2020piqa}, COPA~\cite{roemmele2011choice}, and HellaSwag~\cite{zellers2019hellaswag} assess physical commonsense, causal relationships, and plausible scene continuation; \emph{3) Comprehension \& Logic}: BoolQ~\cite{clark2019boolq} and WinoGrande~\cite{sakaguchi2021winogrande} evaluate boolean reading comprehension and robust pronoun resolution; \emph{4) Language Modeling}: LAMBADA~\cite{paperno2016lambada} measures the model’s proficiency in predicting words based on broad discourse context.

\begin{table*}[tbp]
	\centering
	\caption{\textbf{Zero-shot performance on downstream tasks.} We compare \codename~against the dense baseline by \emph{training from scratch} for 50B tokens. \codename~achieves the best average performance across all the model scales, demonstrating the efficacy of our proposed architecture.}
	\label{tab:benchmark}%
	\resizebox{1.\textwidth}{!}{
		\begin{tabular}{ccc|cccccccccc|c}
			\toprule
			\textbf{Models} &  \makecell{\#~ROM\\Weights} & \makecell{Speed\\(token/s)}  & ARC-E & ARC-C & BoolQ  & COPA & {\footnotesize HellaSwag}  &  {\footnotesize LAMBADA}  & OBQA  & PIQA & SCIQ  & {\footnotesize WinoGrande}     &  Avg. \\
			\midrule
			\multicolumn{3}{c}{\emph{0.6B In-RAM Parameters}\quad\quad\quad\quad} \\
            \midrule
			Baseline & - &  20.1 & 30.5 & 56.0 & 58.3 & 71.0 & 45.7 & 35.5 & \textbf{34.8} & 68.7 & 77.3 & 52.6 & 53.0 \\
			\codename  & 0.5B & 19.9 & \textbf{33.6} & \textbf{60.2} & \textbf{63.0} & \textbf{72.0} & \textbf{49.2} & \textbf{39.8} & 34.6 & \textbf{70.2} & \textbf{78.4} & \textbf{53.8} & \textbf{55.5}  \\
            \midrule
			\multicolumn{3}{c}{\emph{1.7B In-RAM Parameters}\quad\quad\quad\quad} \\
            \midrule
			Baseline & - & 13.8 & 34.4 & 61.7 & 58.9 & 69.0 & 51.7 & 41.3 & 37.6 & 70.8 & 80.6 & 53.6 & 56.0  \\
			\codename   & 1.1B & 13.7 & \textbf{37.9} & \textbf{66.2} & \textbf{62.4} & \textbf{74.0} & \textbf{56.6} & \textbf{45.6} & \textbf{39.0} & \textbf{71.7} & \textbf{85.4} & \textbf{58.7} & \textbf{59.7}  \\
            \midrule
			\multicolumn{3}{c}{\emph{4B In-RAM Parameters}\quad\quad\quad\quad} \\
            \midrule
            Baseline & - &  6.1& 38.0  & 66.3  & 62.7  & \textbf{80.0}  & 57.9  & 45.6  & 39.4  & 72.9  & 84.4  & 58.1  & 60.5  \\
            \codename  & 2.8B & 6.1 & \textbf{42.2}  & \textbf{70.2}  & \textbf{64.4}  & 77.0  & \textbf{62.3}  & \textbf{50.1}  & \textbf{40.2}  & \textbf{75.4}  & \textbf{87.2}  & \textbf{61.6}  & \textbf{63.0}  \\
			\bottomrule
		\end{tabular}%
	}
\end{table*}

\subsection{Experimental Result}

We evaluate the zero-shot performance / accuracy for pre-trained checkpoints on the above benchmarks.

\paragraph{Comparison to Dense Baselines.} \cref{tab:benchmark}~presents a comprehensive comparison between \codename~and dense baselines across diverse benchmarks. The results explicitly validate that separating memory capacity from computational logic yields significant gains without increasing FLOPs.
\begin{itemize}

    \item \emph{Knowledge-Intensive Tasks:} \codename~demonstrates its strongest advantage in tasks requiring factual recall. On ARC-Challenge, MeKi-1.7B achieves a score of 37.9, surpassing the 1.7B baseline (34.4) by +3.5 points and effectively matching the performance of the significantly larger 4B dense model (38.0). Similarly, on SciQ, MeKi-1.7B reaches 85.4, outperforming the baseline by 4.8 points. This suggests that the ROM effectively functions as an extended key-value store for static world knowledge, relieving the FFN parameters from the burden of memorization.

    \item \emph{Reasoning and Context:} In reasoning-heavy tasks like BoolQ and HellaSwag, MeKi-1.7B scores 62.4 and 56.6 respectively, consistently outperforming the baseline (58.9 and 51.7). Notably, on the LAMBADA language modeling benchmark, MeKi-1.7B achieves a score of 45.6, which is identical to the 4B baseline model (45.6). This indicates that the injected "expert vectors" provide crucial semantic anchoring for long-range dependency prediction, effectively simulating the capacity of a 4B parameter model using only 1.7B active parameters and external memory.
\end{itemize}

\begin{figure}[t]
    \centering
    \includegraphics[width=0.45\columnwidth]{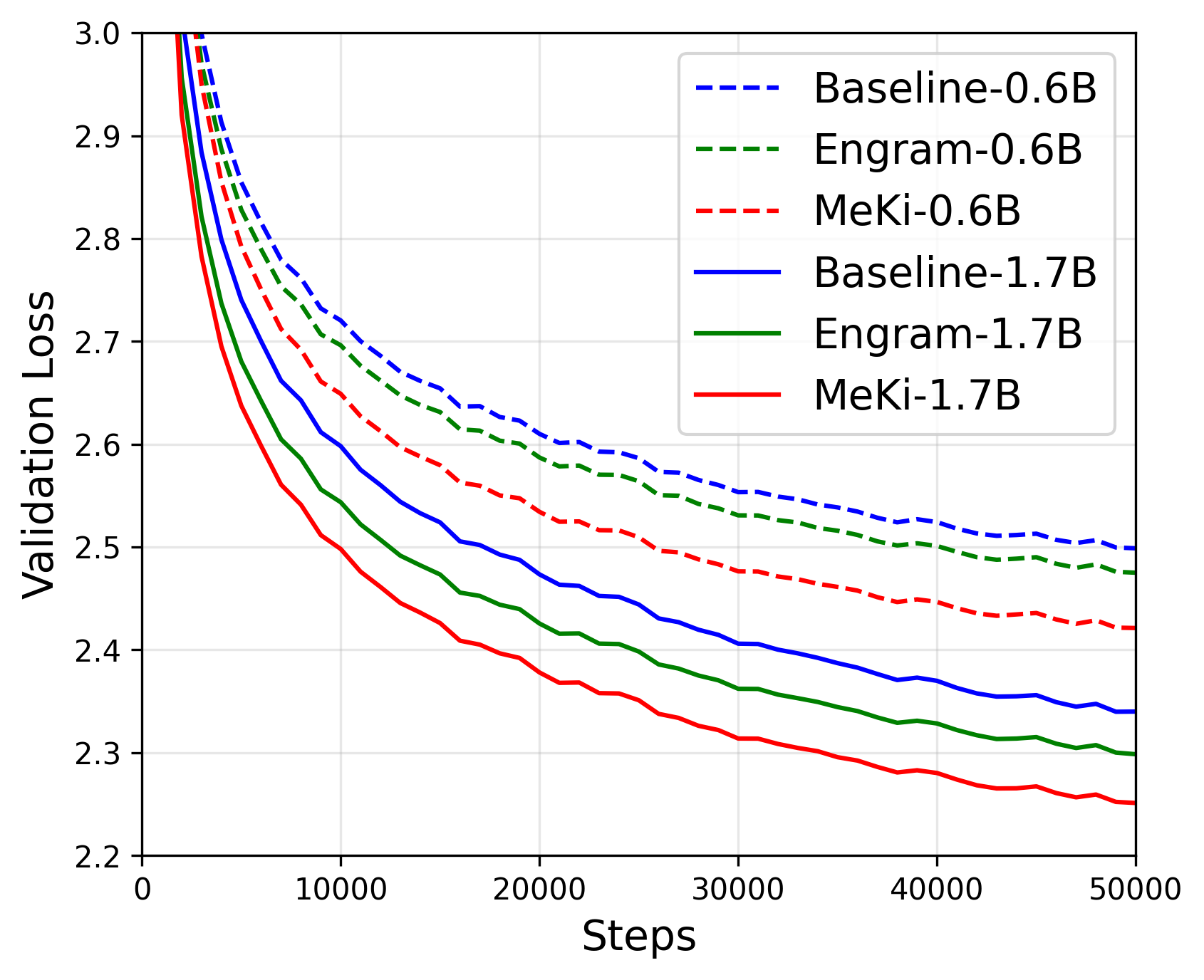}
    \vspace{-3mm}
    \caption{The comparison of validation loss for different methods where \codename~outperforms both baseline and Engram.}
    \label{fig:validation_loss_curve}
\end{figure}

\paragraph{Comparison to Sparse Memory Architectures.} To evaluate the efficacy of our proposed architecture, we compare\footnote{~We implement PLE based on the official code of Gemma-3n and the Engram method based on the official github repository.} \codename~with PLE~\cite{gemma_3n_2025} and Engram~\cite{cheng2026conditional}. As shown in \cref{tab:other_results}, \codename-1.7B attains an average score of 59.7, outperforming PLE and Engram by 2.7 and 1.8 points, respectively. \cref{fig:validation_loss_curve} illustrates the validation loss curves, where \codename~outperforms other approaches by a large margin for both the 0.6B and 1.7B scale. These results demonstrate that \codename~is not merely a storage extension, but a more effective mechanism for integrating large-scale static knowledge into the transformer backbone.

\paragraph{Inference Latency.} In \cref{tab:benchmark}, we report the generation speed (token/s) of our \codename~architecture and its dense baseline counterpart, we use android smartphone with Qualcomm Snapdragon 8 Elite platform and set the KV cache length to be 10K.
During inference, \codename~maintains the same number of active parameters in RAM as the dense baseline. By offloading the memory weights to ROM storage space after re-parameterization, and using asynchronous prefetching directly via token ID, we achieve nearly zero latency overhead.

\begin{table}[tbp]
\centering
\caption{\textbf{Zero-shot performance on downstream tasks.} We compare \codename~against PLE~\cite{gemma_3n_2025} and Engram~\cite{cheng2026conditional} with similar memory size offloaded to ROM. ~*~We re-produce these results, see footnotes on this page for details.}
\resizebox{0.65\columnwidth}{!}{
\begin{tabular}{l|ccc|ccc}
\toprule
\multirow{2}[2]{*}{\textbf{Benchmark}} & \multicolumn{3}{c|}{\textbf{0.6B Scale}} & \multicolumn{3}{c}{\textbf{1.7B Scale}} \\
        & PLE*  & Engram* & {\codename} &  PLE* &  Engram* & {\codename} \\
\midrule
\#RAM Params &  0.61B&    0.60B    &  0.60B& 1.76B &    1.75B   &  1.75B \\
\#ROM Weights &  0.54B &    0.62B   &     0.54B  &  1.09B &   1.24B    &   1.09B\\
\midrule
ARC-C & 31.5  & 31.4  & \textbf{33.6}  &  33.7   &     36.8  & \textbf{37.9} \\
ARC-E      & 58.8  & \textbf{60.3}  & 60.2  &   62.3    &   65.8    & \textbf{66.2}  \\
BoolQ         & 57.9  & 61.9  & \textbf{63.0}  &   61.6    &   60.9    & \textbf{62.4}  \\
COPA          & 67.0  & 68.0  & \textbf{72.0}  &    73.0   &    72.0   & \textbf{74.0}  \\
HellaSwag     & 46.6  & 45.3  & \textbf{49.2}  &    52.7   &    54.0   & \textbf{56.6}  \\
LAMBADA       & 36.1  & 34.5  & \textbf{39.8}  &    41.5   &    42.3   & \textbf{45.6}  \\
OBQA    & \textbf{35.0}  & 33.6  & 34.6  &     36.4  &    36.0  & \textbf{39.0}  \\
PIQA          & 69.1  & 69.2  & \textbf{70.2}  &   69.7    &  \textbf{71.9}     & 71.7  \\
SCIQ          & \textbf{80.3}  & 78.5  & 78.4  &    81.8   &    83.0   & \textbf{85.4}  \\
WinoGrande    & 52.9  & \textbf{54.5}  & 53.8  &   57.4    &   56.3   & \textbf{58.7}  \\
\midrule
Average       & 53.5  & 53.7  & \textbf{55.5}  &    57.0   &    57.9   & \textbf{59.7}  \\
\bottomrule
\end{tabular}%
}
\label{tab:other_results}%
\end{table}%

\section{Ablation Study and Analysis}

In this section, we analyze the critical architectural choices of \codename~to understand their contribution to the overall performance. All the experiments conducted in this section follow the same setting in \cref{sec: setup}.

\begin{table*}[tbp]
\centering
\caption{\textbf{Zero-shot performance with different types of memory.} The two variants are also based on 0.6B baseline model.}
\vspace{-2mm}
\label{tab:component_ablation}
\resizebox{\textwidth}{!}{
\begin{tabular}{l|cc|cccccccccc|c}
\toprule
\textbf{Model} & $ \mathbf{m}^l_{static}$  & $\mathbf{m}^l_{dyn}$ & ARC-E & ARC-C & BoolQ & COPA & {\footnotesize HellaSwag}  &  {\footnotesize LAMBADA}  & OBQA  & PIQA & SCIQ  & {\footnotesize WinoGrande}     &  Avg. \\
\midrule
Baseline-0.6B  &$\times$ &$\times$ & 30.5 & 56.0 & 58.3 & 71.0 & 45.7 & 35.5 & 34.8 & 68.7 & 77.3 & 52.6 & 53.0 \\
Static-Only-0.6B   & \checkmark & $\times$ & 32.8 & 59.7 & 61.4 & 71.0 & 49.0 & 37.4 & \textbf{36.0} & 69.1 & 79.2 & 52.4 & 54.8 \\
Dynamic-Only-0.6B   & $\times$ & \checkmark & 31.2 & 57.5 & 62.0 & 71.0 & 47.7 & 39.0 & 35.6 & 69.3 & \textbf{79.5} & \textbf{54.1} & 54.7 \\
MeKi-0.6B  & \checkmark & \checkmark & \textbf{33.6} & \textbf{60.2} & \textbf{63.0} & \textbf{72.0} & \textbf{49.2} & \textbf{39.8} & 34.6 & \textbf{70.2} & 78.4 & 53.8 & \textbf{55.5} \\
\bottomrule
\end{tabular}
}
\label{tab:memory_type}
\end{table*}

\begin{table*}[tbp]
\centering
\caption{\textbf{Zero-shot performance with \codename~module incorporated at different positions.} All models are based on \codename-0.6B.}
\vspace{-2mm}
\resizebox{\textwidth}{!}{
\begin{tabular}{l|cccccccccc|c}
\toprule
\textbf{Position} & ARC-E & ARC-C & BoolQ  & COPA & {\footnotesize HellaSwag}  &  {\footnotesize LAMBADA}  & OBQA  & PIQA & SCIQ  & {\footnotesize WinoGrande}     &  Avg. \\
\midrule
\circleone~Parallel to FFN &\textbf{33.6} &60.2 &\textbf{63.0} &\textbf{72.0} &\textbf{49.2} &\textbf{39.8} & 34.6 &\textbf{70.2} & 78.4 & 53.8 & \textbf{55.5} \\
\circletwo~Parallel to Attn &31.4 &\textbf{61.0} &61.7 & 71.0 & 48.4 & 38.6 &\textbf{36.0} & 69.0 & 78.4 &\textbf{55.8} & 55.1 \\
\circlethree~After Attn &31.7 &60.2 &58.9 & 71.0 & 48.6 & 37.4 & 35.8 & 69.0 &\textbf{81.7} & 54.9 & 54.9 \\
\circlefour~After FFN &33.3 &60.4 &61.4 & 71.0 & 46.9 & 38.7 & 33.2 & 69.5 & 79.4 & 53.3 & 54.7 \\
\bottomrule
\end{tabular}
}
\label{tab:ablation_position}
\end{table*}

\subsection{Effectiveness of Static and Dynamic Memory}
To verify the individual contribution of the two distinct memory components within \codename~module, we selectively disable the static memory retrieval $\mathbf{m}^l_{static}$ or the dynamic memory projection $\mathbf{m}^l_{dyn}$ during training. We summarize the results in \cref{tab:memory_type} and provides insights below.

\textbf{Static-Only.} Firstly, we evaluate the \emph{Static-Only} variant, which solely relies on the trainable memory embedding table $\mathbf{M}^l$. Therefore \cref{eq:memory_type} reduces to
\begin{equation}
\mathbf{e}^l_t = \alpha^l \cdot \text{RMSNorm}\left( \mathbf{M}^l[x_t]\right).
\end{equation}
Compared to the 0.6B baseline, this variant achieves significant improvement (average score raised from 53.0 to 54.8). This result indicates the static memory effectively learns the token-level priors and factual knowledge directly in the embedding space.

\textbf{Dynamic-Only.} Secondly, we examine the \emph{Dynamic-Only} variant, where the static memory table $\mathbf{M}^l$ is deleted. The token-level expert is derived by applying a non-linear projection $\mathbf{G}^l$ to the global word embeddings. In this situation, \cref{eq:memory_type} reduces to
\begin{equation}
\mathbf{e}^l_t = \alpha^l \cdot \text{RMSNorm}\left(\beta^l \cdot  \mathbf{G}^l(E_{global}[ .x_t])  \right).
\end{equation}
This variant performs similarly to the Static-Only variant. This suggests that complex non-linear transformation is capable of synthesizing expressive layer-specific features from the global semantic space, providing the model with enhanced representability without layer-specific memory.

\codename-0.6B integrates both memory components via the learnable coefficients $\alpha^l$ and $\beta^l$. It surpasses both Static-Only and Dynamic-Only variants across the majority of benchmarks. The consistent gain (+0.7 over Static-Only and +0.8 over Dynamic-Only) suggests that the static memory and dynamic projection capture complementary information. Their combination allows the model to maximize the utilization of the storage budget for knowledge injection.

\subsection{Optimal Position for Memory Injection}
\begin{figure}[t]
    \vspace{-8mm}
    \centering
    \begin{minipage}[b]{0.5\columnwidth}
        \centering
        \includegraphics[width=\linewidth]{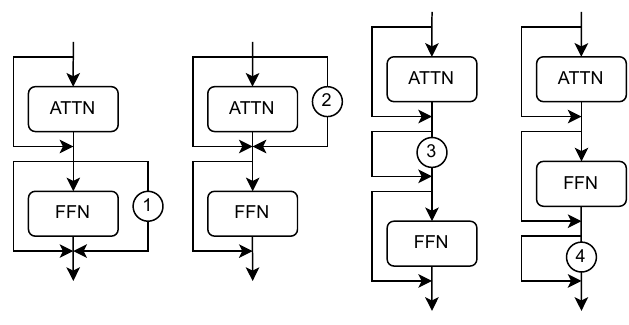}
        \vspace{10mm}
        \label{fig:ablation_position}
    \end{minipage}
    \hspace{10mm}
    \begin{minipage}[b]{0.3\columnwidth}
        \centering
        \includegraphics[width=\linewidth]{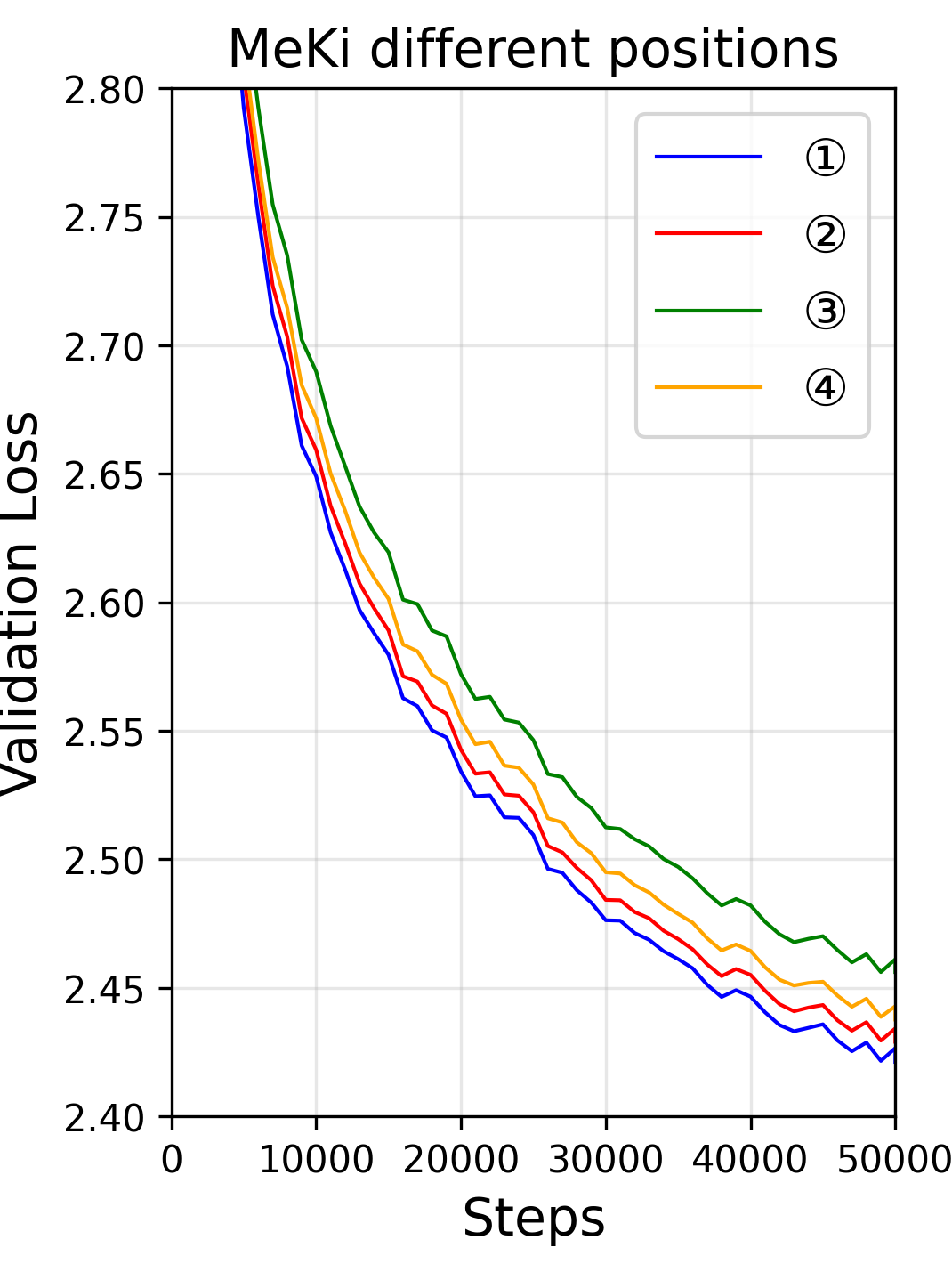}
        \label{fig:ablation_position_loss}
    \end{minipage}
    \vspace{-8mm}
    \caption{(Left) We choose four potential positions to incorporate the \codename~module within our architecture. (Right) Validation loss for different position settings.}
\end{figure}

We investigate the optimal position where the \codename~module should be incorporated within a Transformer layer. As illustrated in \cref{fig:ablation_position}, we compare 4 different placement, \circleone~Parallel to FFN, \circletwo~Parallel to Attention, \circlethree~After Attention, \circlefour~After FFN.

\cref{tab:ablation_position} demonstrates that the option \circleone, Parallel to FFN, achieves the best scores averaged from 10 downstream tasks. \cref{fig:ablation_position_loss} also shows that option \circleone~has the best validation loss during training.
For the option \circlethree~and \circlefour~that achieve the worst performance, we conjecture that the low-rank gating structure inside the \codename~leads to information bottleneck when used alone, causes performance drop. The option \circletwo~is sub-optimal because the Attention is responsible for building the global dependencies among different tokens in a sequence. The token-level expert knowledge at position \circletwo~would play a much weaker role than the situation where it serves as an capacity augmentation parallel to FFN.

\begin{figure}[t]
\centering
\begin{minipage}{0.3\columnwidth}
    \centering
\includegraphics[width=0.99\columnwidth]{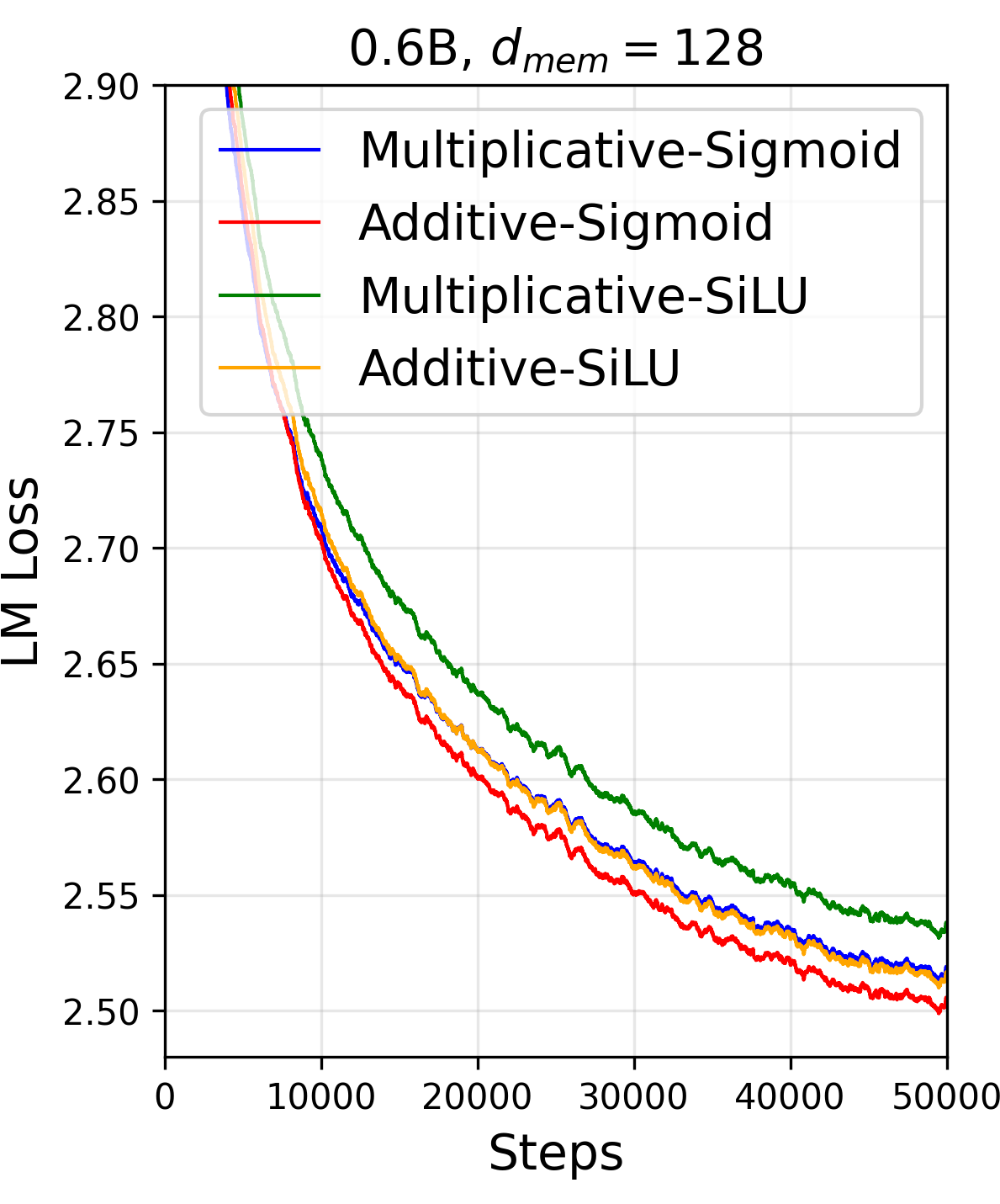}
\end{minipage}
\hspace{10mm}
\begin{minipage}{0.3\columnwidth}
    \centering
\includegraphics[width=0.99\columnwidth]{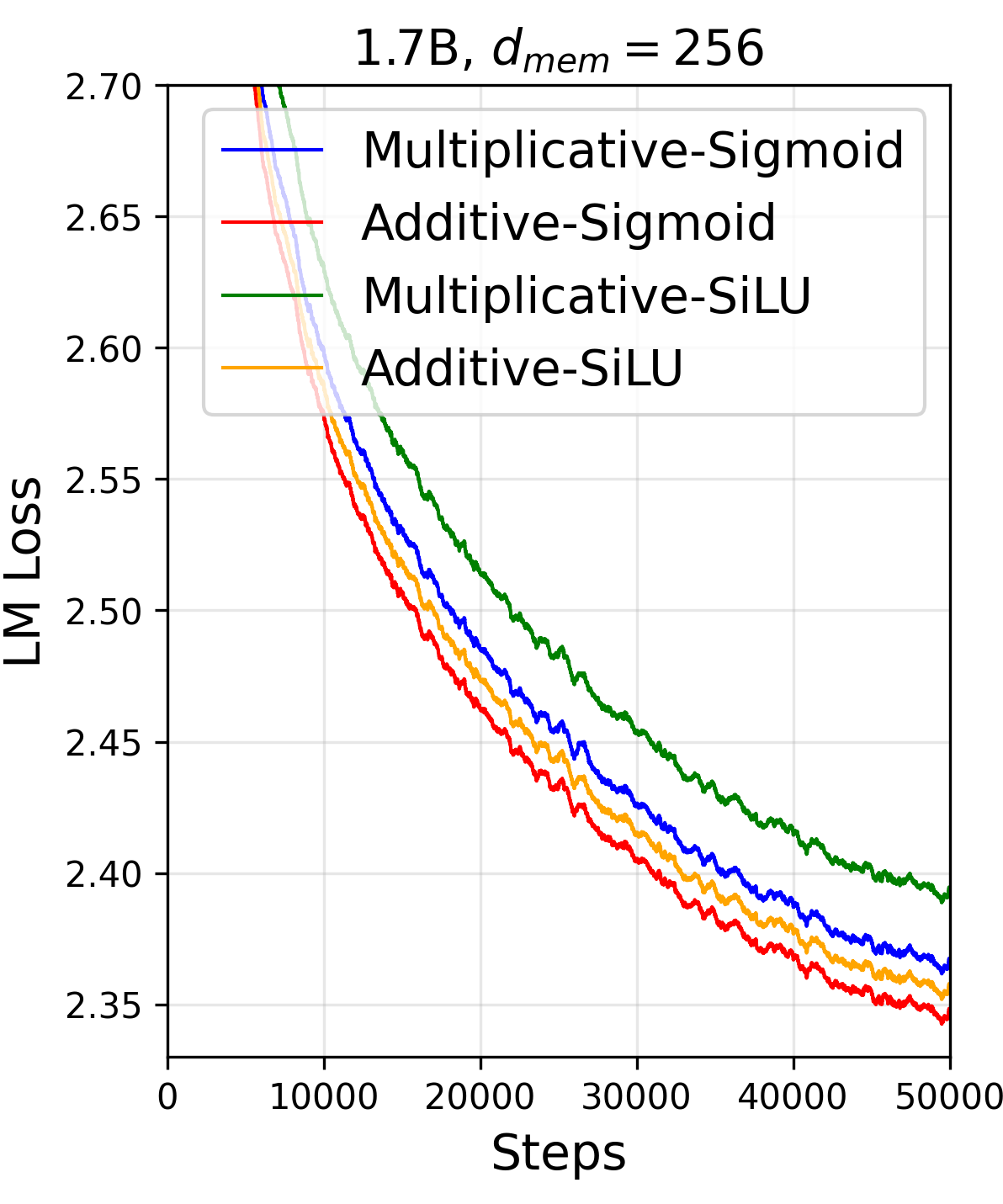}
\end{minipage}
\vspace{-3mm}
\caption{Validation loss over different feature fusion strategies.}
\label{fig:optimal_fuser}
\end{figure}

\subsection{Optimal Feature Fusion}

The integration of the retrieved knowledge vector $\mathbf{e}^l_t$ into the transformer block is critical to maximize the memory expert's utilization. \codename~operate in the low-rank space in order to save FLOPs while injecting expert knowledge $\mathbf{e}^l_t$ into the contextualized token hidden state $\mathbf{h}^l_t$:
\begin{equation}
\mathbf{v}^l_t = \mathbf{e}^l_t + \mathbf{g}^l_t = \mathbf{e}^l_t + \sigma(\mathbf{W}^{l}_{gate} \mathbf{h}^l_t).
\end{equation}
In total, we explored four different feature fusion strategies as follows. We omit subscript and superscript for simplicity.
\begin{itemize}
\vspace{-2mm}
    \item Additive-Sigmoid: $\mathbf{v} = \mathbf{e} + \sigma(\mathbf{W}_{gate} \mathbf{h}),$
    \item Multiplicative-Sigmoid: $\mathbf{v} = \mathbf{e}  \odot \sigma(\mathbf{W}_{gate} \mathbf{h}),$
    \item Additive-SiLU: $\mathbf{v} = \mathbf{e} + \text{SiLU}(\mathbf{W}_{gate} \mathbf{h}),$
    \item Multiplicative-SiLU: $\mathbf{v} = \mathbf{e} \odot \text{SiLU}(\mathbf{W}_{gate} \mathbf{h}).$
\vspace{-2mm}
\end{itemize}

We train all these variants from scratch based on \codename-0.6B and \codename-1.7B, and show the corresponding training loss curves in \cref{fig:optimal_fuser}. As demonstrated in the loss curves, the ``Additive-Sigmoid" method achieves the best performance among these fusion strategies.

\subsection{Scaling Law of Memory Size}
\vspace{-1mm}
We investigate the scaling properties of the proposed \codename~architecture by explore the impact of the memory dimension $d_{mem}$ used by the knowledge vectors. For \codename-0.6B, we vary the $d_{mem}$ in [64, 96, 128, 160, 192], which directly leads to the memory size of \codename-0.6B being [$2.72\times 10^8$, $4.08\times 10^8$, $5.44\times 10^8$, $6.80\times 10^8$, $8.15\times 10^8$] respectively. We compute the number of Memory Weights according to $L\times |\mathcal{V}| \times d_{mem}$, where $L$ is the number of transformer layers. Similarly for \codename-1.7B, we vary the $d_{mem}$ in [192, 256, 320, 384, 448], which leads to the memory size of \codename-1.7B being [$8.15\times 10^8$, $1.09\times 10^9$, $1.36\times 10^9$, $1.63\times 10^9$, $1.90\times 10^9$] respectively.

\begin{figure}[t]
\centering
\begin{minipage}{0.49\columnwidth}
    \centering
    \includegraphics[width=\columnwidth]{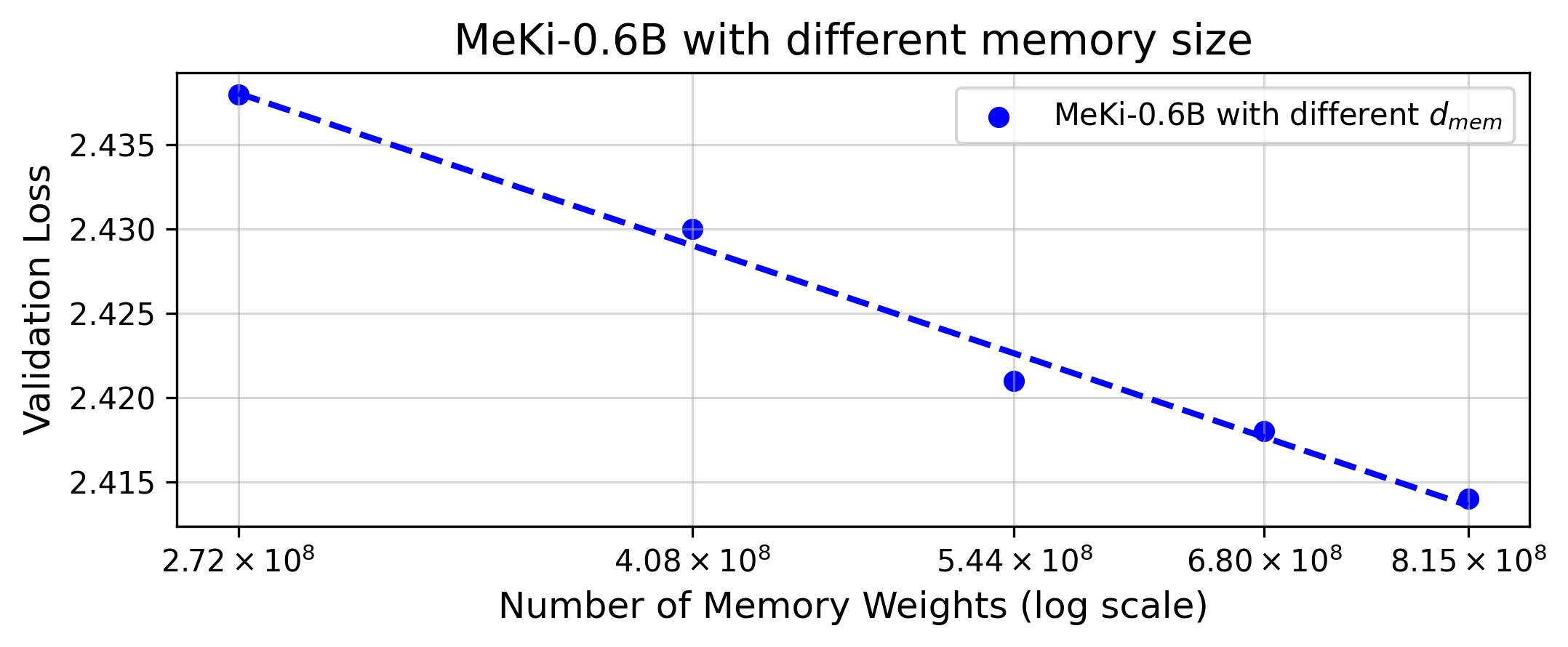}
\end{minipage}
\hfill
\begin{minipage}{0.49\columnwidth}
    \centering
    \includegraphics[width=\columnwidth]{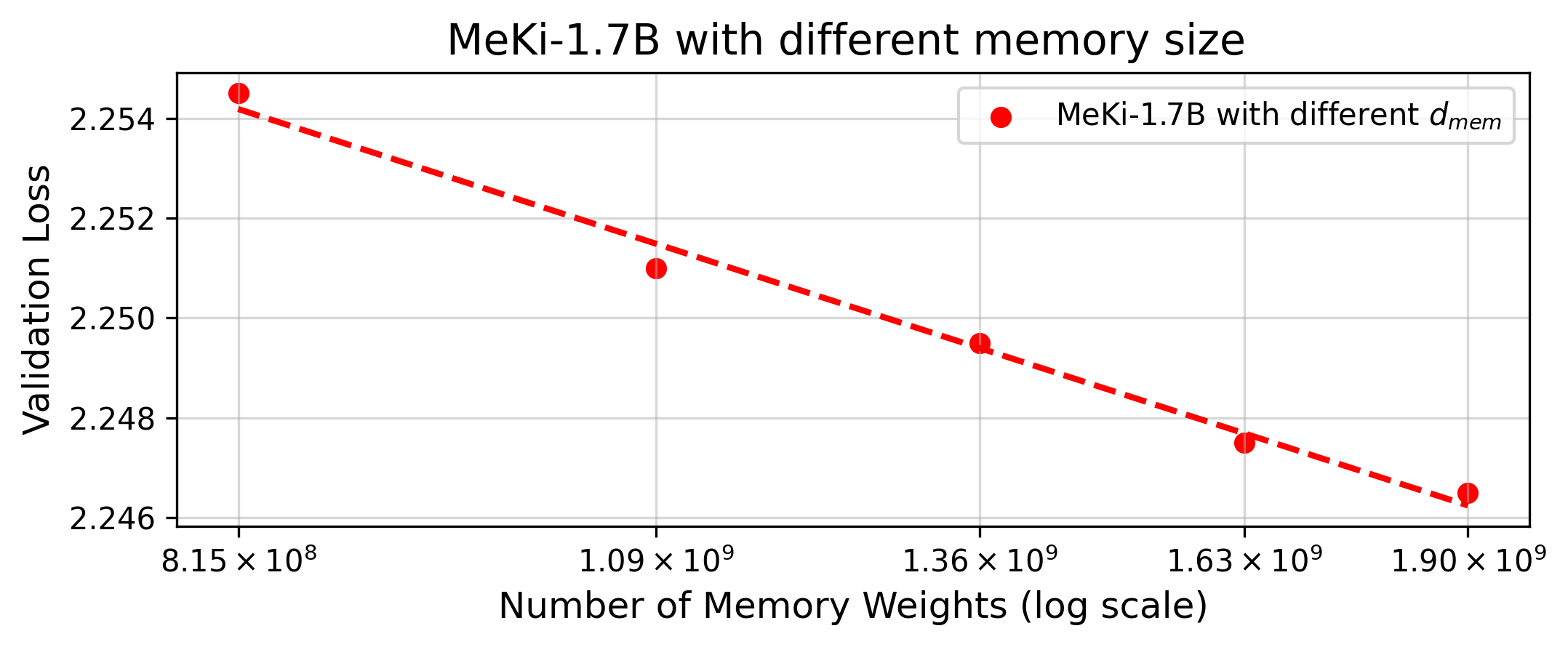}
\end{minipage}
\vspace{-2mm}
\caption{The final validation loss values with different memory sizes. All the models are trained from scratch for 50 billion tokens.}
\vspace{-6pt}
\label{fig:scaling_law}
\end{figure}

\begin{figure}[t]
    \centering
    \begin{minipage}{0.495\columnwidth}
        \centering
        \includegraphics[width=\columnwidth]{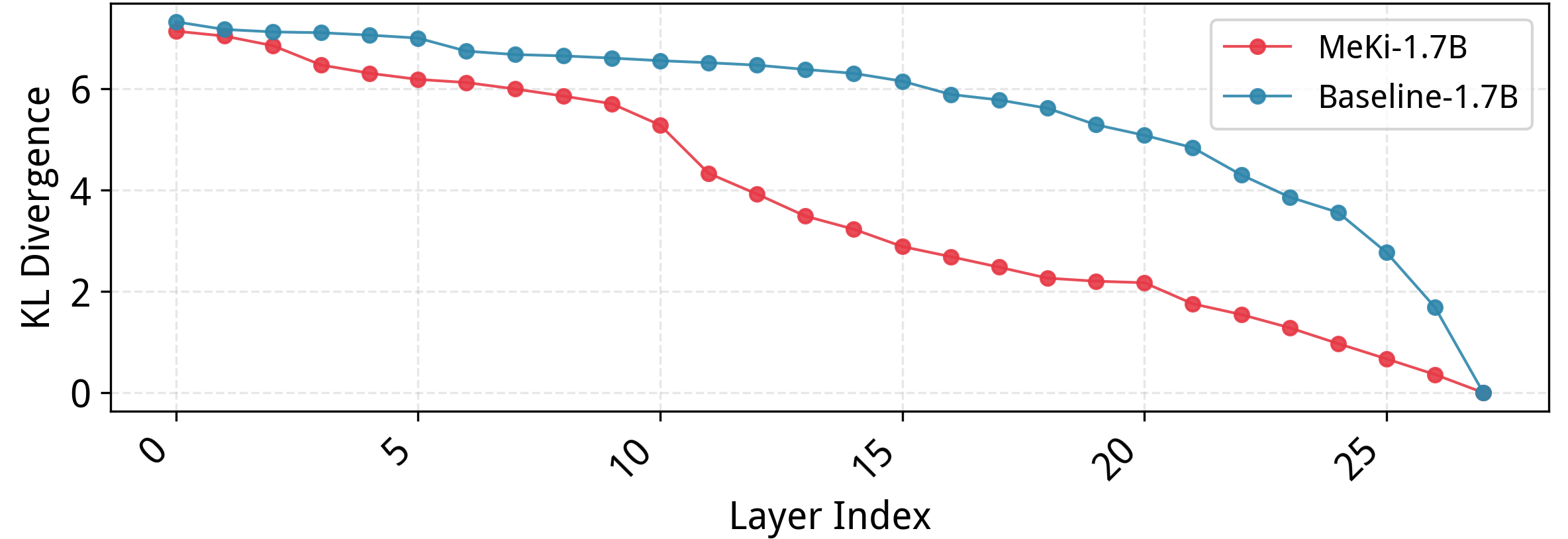}
    \end{minipage}
    \hfill
    \begin{minipage}{0.495\columnwidth}
        \centering
        \includegraphics[width=\columnwidth]{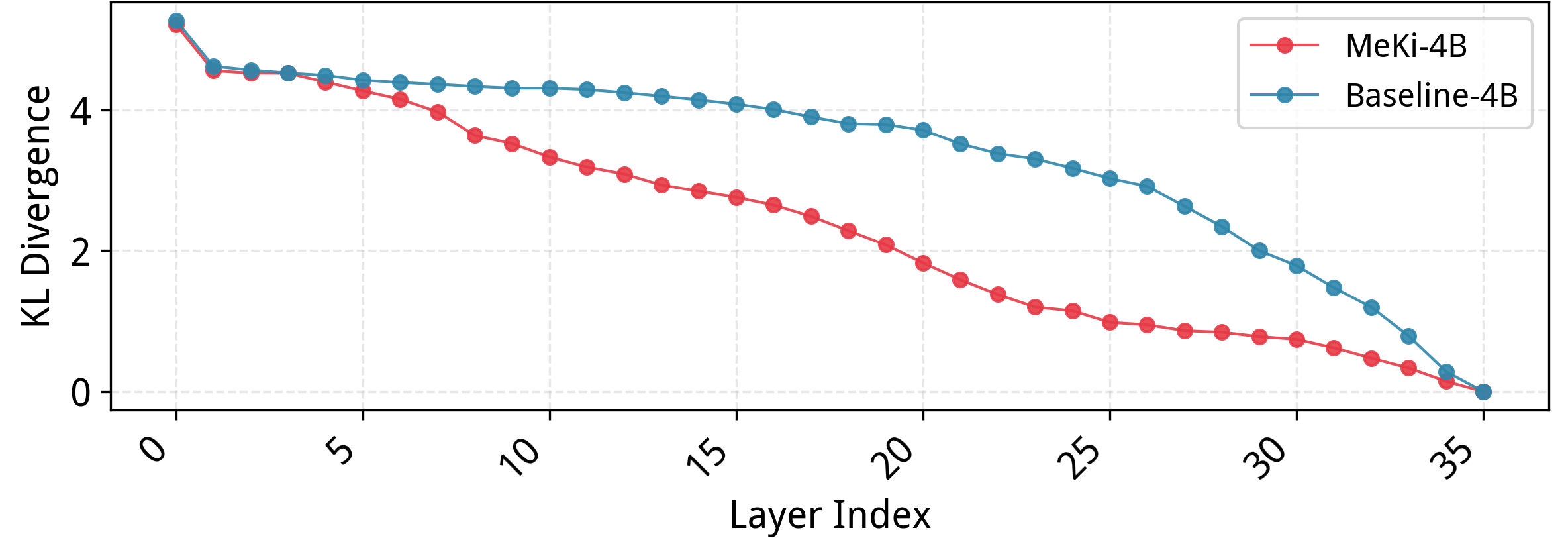}
    \end{minipage}
    \vspace{-1mm}
    \caption{Layer-wise KL Divergence Analysis. \codename~exhibits a lower divergence across all layers, demonstrating its ability to accelerate prediction convergence.}
    \label{fig:kl}
    \end{figure}

We illustrate the final validation loss of different models under the same training data budget and the fixed active parameter budget in \cref{fig:scaling_law}.
As anticipated, we observe that the model performance consistently exhibits a \textbf{log-linear} trend with respect to the memory size. In order to balance the storage cost (\eg~ROM space) and model capacity, we select the dimension $d_{mem}=128$ for \codename-0.6B and $d_{mem}=256$ for \codename-1.7B as the optimal trade-off for our main experiments.

\subsection{Accelerating Prediction Convergence}
\vspace{-1mm}
We employ the LogitLens~\cite{nostalgebraist2020logitlens} to examine how predictions evolve across the model's layers. We input the intermediate hidden states of each layer into the final classifier head, we calculate the KL divergence between these early predicted and the final output distribution. This measures  convergence of latent representations toward their final predicted state.

\cref{fig:kl} reports the statistics by averaging the results of 10K tokens sampled from validation set. Compared to the baseline, MeKi exhibits a systematically lower KL divergence across all layers. The consistently lower values indicate that MeKi facilitates a more rapid alignment of intermediate feature distributions with the final output. This observation suggests that by retrieving knowledge vector from the layer-specific memory bank, MeKi accelerates prediction convergence, enabling the model to reach high-confidence states earlier inside the network architecture.

\subsection{Scaling Training-Time FLOPs}

During training, the projector $\mathbf{G}^l(\cdot)$ dynamically maps the global word embeddings to the \codename's layerwise lower-dimensional knowledge space. We try to use a simple linear projection layer as $\mathbf{G}^l$ to replace the current SwiGLU function. We observe degradation in both the training loss curve and the downstream performance across all model sizes. During the inference stage, no matter how the projection $\mathbf{G}^l$ is modeled, it will be absorbed into the final embedding memory $\tilde{\mathbf{M}}^l$ by reparameterization techniques, leading to no computation overhead at all. This suggests that increasing the FLOPs of the training phase is still helpful to the final performance even though these FLOPs are never used for inference.

\section{Conclusion}

\codename~introduces a new memory-based paradigm for scaling LLM, decoupling model capacity from computational cost by leveraging abundant ROM space to circumvent the limitations of on-device resources constraints. By re-parameterizing the computational overhead of training into static memory lookup tables, MEKI achieves zero inference latency overhead while effectively expanding on-device model capacity via storage. Extensive experiments demonstrate that our approach enables a 1.7B model to rival the performance of a 4B dense model while maintaining identical decoding speed on Qualcomm Snapdragon hardware. Our findings validate that shifting the scaling bottleneck from computation to memory is a highly effective strategy for powerful on-device AI.


\bibliography{colm2026_conference}
\bibliographystyle{colm2026_conference}

\newpage
\appendix
\onecolumn
\section{Appendices}

\subsection{Detailed Model Architecture and Hyper Parameters}

\begin{table}[h]
\centering
\caption{Detailed model architecture information and training hyper parameters.}
\label{tab:hyper-llama-opt}
\begin{tabular}{l|ccc}
\toprule
\bf Hyper-Parameters & \bf \codename-0.6B & \bf \codename-1.7B  & \bf \codename-4B \\
\midrule
\# RAM Params &  0.60B& 1.75B & 4.12B \\
\# ROM Weights & 0.54B &  1.09B& 2.80B \\
\# Training Tokens &50B &50B &50B  \\
\midrule
Layers & 28 & 28 & 36  \\
Hidden Size  $d_{model}$ & 1024 & 2048 &  2560 \\
\codename~Dim $d_{mem}$ & 128 &  256 &  512 \\
Intermediate Size & 3072 & 6144 &  9728 \\
\midrule
Attention module & \multicolumn{3}{c}{Group Query Attention} \\
RoPE $\theta$ & \multicolumn{3}{c}{500000}  \\
Training Steps $\theta$ & \multicolumn{3}{c}{50000}  \\
Sequence Length  & \multicolumn{3}{c}{4096}  \\
Vocab Size  & \multicolumn{3}{c}{151680}  \\
Batch Size  & \multicolumn{3}{c}{256}  \\
Base Learning Rate  & \multicolumn{3}{c}{4e-4}  \\
Lr Scheduler & \multicolumn{3}{c}{Cosine}  \\
Adam $\beta$ & \multicolumn{3}{c}{(0.9, 0.95)}  \\
Weight Decay & \multicolumn{3}{c}{0.1}  \\
Gradient Clip & \multicolumn{3}{c}{1.0}  \\
\bottomrule
\end{tabular}
\end{table}

\subsection{Detailed Statistics for NPU Inference}

\begin{figure*}[htbp]
\centering
\vspace{0.3cm}
\includegraphics[width=0.6\columnwidth]{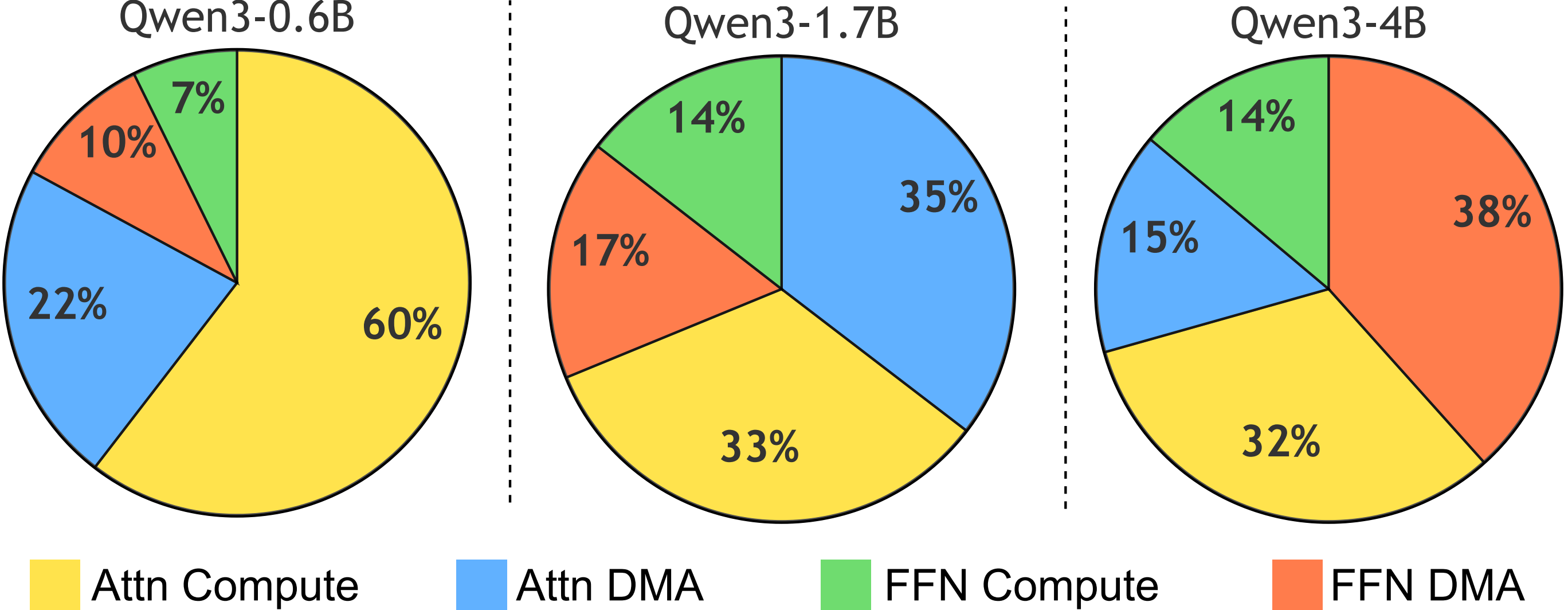}
\caption{The percentage of time consumption of different model components during NPU inference.}
\label{fig:npu_time_breakdown}
\end{figure*}

For small-sized Qwen3 model family, we measure the percentage of time consumption for different model components during the on-device inference process. We get the statistics by measuring the generation speed (token/s) on Qualcomm Snapdragon 8 Elite with KV cache length being 10K. 

The \textbf{\emph{DMA}} part refers to the time spent on moving the corresponding parameters from RAM to NPU. During inference, the Qwen3-4B model spends 38\% of its time moving FFN parameters from RAM to NPU. Besides, time spent on FFN computation is actually less than the parameter movement, where 32\% of the inference time is used for FFN computation. The DMA time becomes a major bottleneck why scaling the number of parameters of dense LLM makes inference extremely slower on NPU. Deploying MoE architecture requires much more DMA time than dense LLM.

\end{document}